\documentclass[aps,prl,twocolumn,groupedaddress]{revtex4-2}
\usepackage{amsmath,amssymb,amsthm,mathrsfs,amsfonts,dsfont} 
\usepackage{bm,booktabs,enumitem,placeins}
\usepackage{graphicx}
\usepackage{subfigure}
\usepackage{amsbsy}
\usepackage{bm}
\usepackage{orcidlink}
\usepackage{xcolor}

\usepackage{algorithm}
\usepackage{algpseudocode}
\usepackage{algorithmicx}

\newcommand{\avg}[1]{\langle #1\rangle}
\newcommand{\E}{\mathbb{E}}

\newcommand{\tr}{\operatorname{tr}}
\newcommand{\diag}{\operatorname{diag}}

\def\a{\alpha}

\begin{document}

\title{Exact theory of next-token prediction}
\author{Mengyao An$^{1}$}
\author{Haiping Huang\,\orcidlink{0000-0001-8757-4733}$^{1,2}$}
\email{huanghp7@mail.sysu.edu.cn}
\affiliation{$^{1}$PMI Lab, School of Physics, Sun Yat-sen University, Guangzhou 510275, People's Republic of China}
\affiliation{$^{2}$Guangdong Provincial Key Laboratory of Magnetoelectric Physics and Devices, Sun Yat-sen University, Guangzhou 510275, People's Republic of China}
\date{\today}

\begin{abstract}
	The next-token prediction ability of self-attention-based transformer is studied by using statistical mechanics and random matrix theory. Our study constructs an exact theory of free energy, state equation, and test-error scaling behavior in both finite and infinite context-length settings. In particular, a double descent phenomenon is revealed for zero-temperature learning, suggesting that the data load and the context length are coupled. In the presence of infinite context length, the errors are exactly those of deterministic ridge shrinkage. Theory predicts all single-instance learning of finite but large token dimension.
\end{abstract}

 \maketitle

\textit{Introduction.}---
Self-attention-based transformer structure is the foundation of large language models and other artificial intelligence foundation models~\cite{Att-2017}. After network parameters are learned in an end-to-end fashion together with post-training (e.g., by supervised fine-tuning and reinforcement learning from human feedback), these complex networks display emergent reasoning capabilities in complex problems such as proving mathematical hypotheses~\cite{Sparks-2023,AGI-2026, Nature-2026}. The learning relies on a single training principle, i.e., autoregressive next-token prediction. Although this principle is completely different from the biological one, the underlying mechanics of next-token prediction remain obscure and mysterious~\cite{Huang-2024,Uni-2023,Li-2024}. 

In recent years, statistical mechanics has played an important role in revealing working principles of in-context learning~\cite{Aky-2022,Ganguli-2024,Li-2025,Bai-2026} and attention-indexed models~\cite{Lenka-2025,Elvis-2026}. We follow this route and construct the statistical mechanics model of next-token prediction in self-attention-based transformer networks. Although the sequence data is complex and non-linear in affecting learning, the model has an exact mean-field solution where the replica symmetry ansatz is stable and coincides with random matrix theory. We derive test error formulae in both Gibbs and Bayes settings, discuss the asymptotic scaling in Bayes optimal conditions, and moreover uncover the double descent phenomenon for next-token prediction learning. All experimental observations about learning are perfectly predicted by our theory.

\textit{Model.}---
We consider a single-head linear self-attention model with data matrix $X^\mu\in\mathbb R^{D\times N}$, $\mu=1,\dots,P$, which has iid entries $X^\mu_{ia}\sim\mathcal N(0,\sigma^2)$ with $\sigma^2=1/D$.
 Write $x^\mu_a$ for its $a$-th column (token). The transformer output is
\begin{equation}
\begin{aligned}
Y^\mu&=\frac1N X^\mu (X^\mu)^\top W X^\mu=S_\mu WX^\mu,\\
S_\mu&\equiv\frac1NX^\mu(X^\mu)^\top\in\mathbb R^{D\times D}.
\end{aligned}
\end{equation}
Only the last column $y^\mu\equiv Y^\mu_{:,N}=S_\mu Wx^\mu_N$ is learned from a teacher model. $N$ is named context length, while $D$ is the token's dimension. The student $W=\diag(w)$ and the teacher $W^*=\diag(w^*)$ are diagonal, with $w^*_i\sim\mathcal N(0,1)$ iid. For any vector $x$,
$Wx=\diag(x)\,w$. Hence,
\begin{equation}
y^\mu=A_\mu w,\qquad y^{*\mu}=A_\mu w^*,
\label{eq:yAw}
\end{equation}
where $A_\mu=S_\mu\diag(x_N^\mu)$. This is the classic teacher-student model in statistical mechanics of learning~\cite{Gyo-1990,Haim-1992}.

The Hamiltonian is written as
\begin{equation}
\mathcal H(w)=\frac{1}{2P}\sum_{\mu=1}^P\lVert y^\mu-y^{*\mu}\rVert^2+\frac{\lambda}{2\beta}\lVert w\rVert^2 .
\end{equation}
$\beta$ is the inverse temperature, while $\lambda$ is a regularization parameter. With $\delta\equiv w-w^*$, each residual is $y^\mu-y^{*\mu}=A_\mu\delta$, so
\begin{equation}
\mathcal H(w)=\frac12(w-w^*)^\top J\,(w-w^*)+\frac{\lambda}{2\beta}\,w^\top w,\\
\label{eq:Hquad}
\end{equation}
where $J$ is a $D\times D$ positive semi-definite empirical Gram matrix of the data, defined as $J\equiv\frac1P\sum_{\mu=1}^PA_\mu^\top A_\mu$. The Boltzmann distribution is thus given by $\frac{1}{Z}e^{-\beta\mathcal{H}}$, where $Z$ is the partition function in physics. To keep the two terms in $\beta\mathcal{H}$ the same order of $D$, $\beta=\frac{bN^2D^2}{N+1}$ for a finite $N$, while $\beta=\frac{b\nu D^3}{1+\nu}$ for extensive $N=\nu D$~\cite{SM}. We consider the thermodynamic limit $D\to\infty$ in this paper. The number of training examples is given by $P=\a D$. In the following, we first study the finite context length and briefly comment on the extensive length at the end.

 \textit{Random matrix theory.}---
Before taking the disorder average, we carry out the Gaussian integral to get
\begin{equation}
\ln Z=\frac D2\ln2\pi-\frac12\ln\det H-\frac\lambda2\,w^{*\top}\big(I-\lambda H^{-1}\big)w^*,
\label{eq:lnZ}
\end{equation}
where $H\equiv\beta J+\lambda I_D$. Thus the posterior mean is a ridge estimator $\langle w\rangle=(J+\tfrac\lambda\beta)^{-1}Jw^*$. The free energy density $f=-\beta^{-1}\ln Z/D$, and in $D\to\infty$, $f$ concentrates on its quenched average $\bar f=\E f$.

Taking the disorder ($X$ and $w^*$) average, we get 
\begin{equation}
-\beta\bar f=\frac12\ln2\pi-\frac1{2D}\E\ln\det H-\frac\lambda2\big(1-\lambda\chi\big),
\label{eq:fbar}
\end{equation}
where $\chi(\lambda)\equiv\frac1D\E\tr H^{-1}=\frac1\beta\,s_J\!\left(-\frac\lambda\beta\right)$ and $\frac{\partial}{\partial\lambda}\frac1D\E\ln\det H=\chi$. Note that $s_J(z)=\frac1D\E\tr(J-z)^{-1}$ is exactly the Stieltjes transform of the spectral law of $J$.

Setting $\theta\equiv\frac{\beta}{PN^2D}\simeq\frac{b}{\a(N+1)}$, one can derive the closed equation for $\chi$:
\begin{equation}\label{eq:chi}
\frac1\chi=\lambda+\alpha\sum_{c\in\{1,2\}}m_c\,\frac{c\theta}{1+c\theta\chi},
\end{equation}
where multiplicities $m_1=N-1$, and $m_2=1$. Because $\frac{\partial}{\partial\lambda}\frac1D\E\ln\det H=\chi$, an integration leads to 
\begin{equation}
\frac1D\E\ln\det H=-\ln\chi-(1-\lambda\chi)+\alpha\sum_cm_c\ln(1+\theta c\chi),
\label{eq:logdet}
\end{equation}
satisfying the boundary condition $\frac1D\ln\det H\to\ln\lambda$ as $\lambda\to\infty$. Technical details are devoted to SM~\cite{SM}.

Collecting all the above results, we arrive at the disorder-averaged free energy density:
\begin{equation}
-\beta\bar f=\frac12\ln(2\pi\chi)+\frac{1-\lambda\chi}{2}-\frac\alpha2\sum_{c}m_c\ln(1+\theta c\chi)-\frac\lambda2\big(1-\lambda\chi\big),
\label{eq:ffinal}
\end{equation}
where $\chi$ the root of Eq.~\eqref{eq:chi}. 

\textit{Replica theory.}---
Applying replica tricks from disorder system theory, one can also compute the disorder-averaged free energy density:
$\E\ln Z=\lim_{n\to0}\frac{\partial}{\partial n}\ln\E Z^n$~\cite{Huang-2022}, where
\begin{equation}
\E Z^n=\int dQ\,d\hat Q\;e^{D\,S_n(Q,\hat Q)},
\end{equation}
where
\begin{equation}
\begin{split}
S_n&=\underbrace{-\frac12\sum_{r,s}\hat Q_{rs}Q_{rs}}_{\text{coupling}}
\;\underbrace{-\frac\alpha2\sum_cm_c\ln\det(I_n+\theta cQ)}_{\text{energetic}}\\
&+\underbrace{\ln\E_{w^*}\!\!\int\!\prod_{r=1}^ndw^r\,e^{-\frac\lambda2\sum_r(w^r)^2+\frac12\sum_{r,s}\hat Q_{rs}(w^r-w^*)(w^s-w^*)}}_{\text{entropic }(\equiv\,\ln\mathcal Z_{\rm site})},
\end{split}
\label{eq:action}
\end{equation}
where $Q_{rs}=\delta^r\cdot\delta^s/D$, and $\delta^r=w^r-w^*$ as before. Next, we adopt the replica-symmetric (RS) ansatz:
\begin{equation}
\begin{aligned}
Q_{rr}&=Q_0,\quad Q_{r\ne s}=Q_1,\\
 \hat Q_{rr}&=\hat Q_0,\quad \hat Q_{r\ne s}=\hat Q_1,\\
\end{aligned}
\end{equation}
from which $\chi\equiv Q_0-Q_1$, which is the posterior variance. One can also get the Gibbs error $\frac1D\E\langle\lVert\delta\rVert^2\rangle=Q_0$, and the error of the posterior mean $Q_1=\frac1D\E\lVert\langle\delta\rangle\rVert^2$.

Using the Laplace method, the saddle-point equation is exactly the same as Eq.~\eqref{eq:chi}, and $Q_1=\frac{\lambda^2}{D}\E\tr H^{-2}=-\lambda^2\partial_\lambda\chi=\frac{\lambda^2\chi^2}{1-\alpha\chi^2g_2}$, where $g_2\equiv\sum_cm_c\Big(\frac{c\theta }{1+c\theta\chi}\Big)^2$. We show in~\cite{SM} that the RS solution is stable everywhere due to the inequality $\a\chi^2g_2<1$, and no replica symmetry breaking occurs. Our theory of next-token prediction is exact in this sense.

\textit{Observables.}---
We next define several observables to measure the learning performance. First, the Gibbs test error $\varepsilon_g^{\rm Gibbs}=\E\Big\langle\lVert A_{\rm new}(w-w^*)\rVert^2\Big\rangle/a_0=Q_0$, where $a_0\equiv\E\lVert Aw^*\rVert^2$ is the signal scale (the dominant one is taken~\cite{SM}), $\E$ defines the disorder average, while $\langle\cdot\rangle$ denotes the thermal average.
Second, the Bayes test error reads $\varepsilon_g^{\rm Bayes}=\E\lVert A_{\rm new}(\langle w\rangle-w^*)\rVert^2/a_0=Q_1$. Thus $\varepsilon_g^{\rm Gibbs}\ge\varepsilon_g^{\rm Bayes}$. One can also derive the expected training error from both random matrix and replica theory that $\varepsilon_t^{\rm Gibbs}=\E\avg{\delta^\top J\delta}/a_0=\frac{1}{b}\Big[\,1-\lambda\chi+\lambda^2\chi-\lambda Q_1\Big]$. Details are given in~\cite{SM}.

Third, we consider Bayesian optimal learning. The prior and likelihood should both be matched~\cite{Huang-2022}. Hence,  $\lambda=1$. In the case of noiseless teacher, $\theta\to\infty$, we have $\chi=Q_1=\operatorname{max}(0,1-\a N)$, and $Q_0=2Q_1$. The minimum mean-squared error is $1-\alpha N$, and perfect recovery occurs at $P=D/N$. Considering the noisy teacher: $y^{*\mu}=A_\mu w^*+\xi^\mu$ with $\xi^\mu_i\sim\mathcal N(0,\Delta)$, we have the Bayes-optimal point: $\lambda=1$, and $\beta=P/\Delta$, where $\Delta=\hat{\Delta}/(N^2D)$ since $\theta=\frac{1}{\Delta N^2D}$. Hence $\hat{\Delta}\sim O(1)$. The posterior mean acquires a noise term: $\langle w\rangle=H^{-1}\big(\beta Jw^*+\beta v\big)$ with $v=\frac1P\sum_\mu A_\mu^\top\xi^\mu$, $\E v=0$, $\operatorname{Cov}v=\frac\Delta PJ$. Hence, $Q_1=Q_1^{(0)}+\tau\Big(\chi-\frac{Q_1^{(0)}}{\lambda}\Big)$, where $\tau=\frac{\beta\Delta}{P}$, $Q_1^{(0)}$ is the counterpart of $\Delta=0$. At the Bayes-optimal point, we get the Nishimori identity: $Q_1=\chi$, $Q_0=\chi+Q_1=2Q_1$, and $\varepsilon_g^{\rm Gibbs}=2\,\varepsilon_g^{\rm Bayes}$, showing that the teacher is statistically indistinguishable from a posterior sample~\cite{Nishimori-2001}.

\textit{Test-error scaling law.}---
We then study the asymptotic limit of test errors. First, $\chi$ is expanded in the large $\a$ limit~\cite{SM}: $\chi\simeq\chi_0+\rho\,b^2\chi_0^3$, and $Q_1^{(0)}\simeq\lambda^2\chi_0^2+3\rho\,\lambda^2b^2\chi_0^4$, where $\chi_0=\frac1{\lambda+b}$, $\rho$ is the relative spectral variance of the bulk of $J$ and its value is given by $\rho=\frac{N+3}{\alpha(N+1)^2}$ for finite $N$ but $\frac{\zeta}{P}$ for extensive $N$~\cite{SM}. $\zeta=2+\frac{1+4\nu}{\nu(1+\nu)}$. Hence, we get in the Bayes-optimal setting $\varepsilon_g^{\rm Bayes}\simeq\sigma(1+\rho)-\sigma^2+O(\alpha^{-3})$, where $\sigma=\frac{\hat\Delta}{\a(N+1)}$.

Taking the ridgeless limit $\lambda\to0$, the state equation [see Eq.~\eqref{eq:chi}] becomes $\Upsilon(\chi)\equiv\alpha\sum_cm_c\frac{c\theta\chi}{1+c\theta\chi}=1$. $\Upsilon$ increases strictly from $\Upsilon(0)=0$ to $\Upsilon(\infty)=\alpha N$, so a finite positive root exists if and only if $\alpha N>1$. Hence, $Q_1^{0}\to0$ and $Q_1^{(0)}/\lambda\to0$, and then $\varepsilon_g^{\rm Gibbs}=(1+\tau)\chi$. In the large $\a$ limit, $\chi\simeq\frac{1+\rho}{b}$. It then follows that $\varepsilon_g^{\rm Bayes}=\tau\chi\simeq\sigma(1+\rho)$, and $\varepsilon_g^{\rm Gibbs}\simeq\frac{(1+\tau)(1+\rho)}{b}$. 

\begin{figure*}
\includegraphics[width=\textwidth]{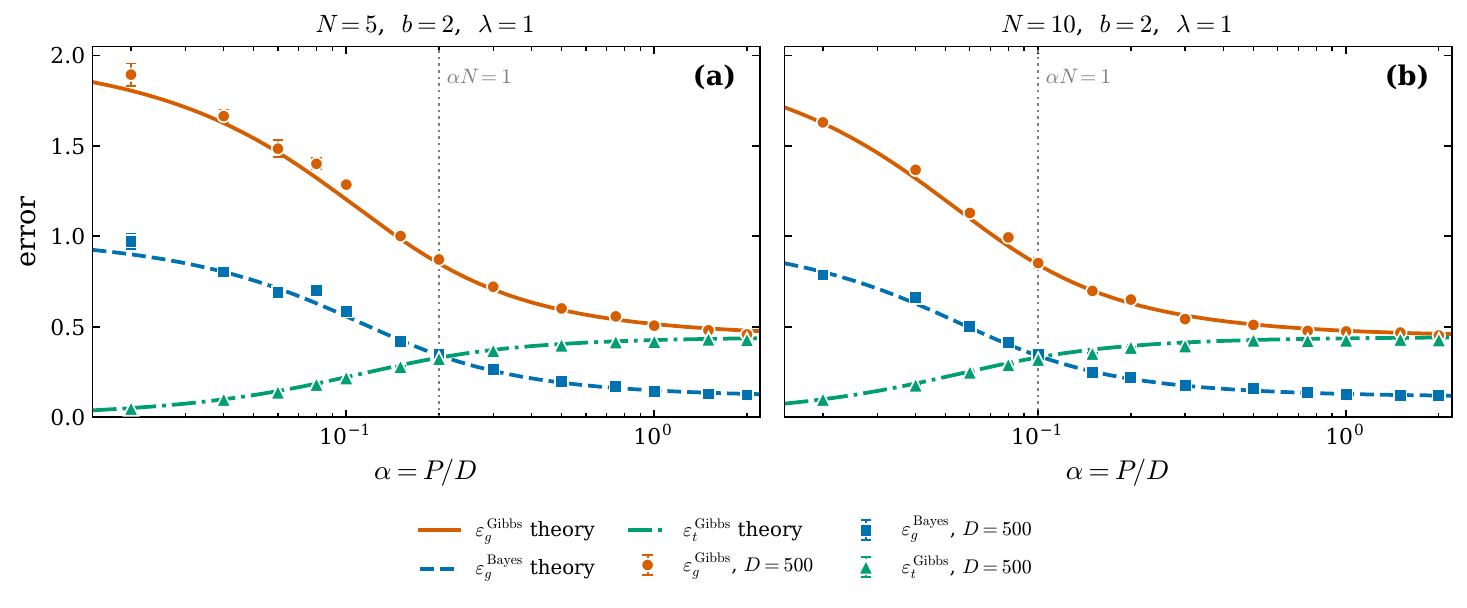}
\caption{\label{fig01}
Bayes and Gibbs error for noiseless learning. We consider two finite context lengths ($N=5$, $10$). Theory is compared with experiments on single learning instances of $D=500$. 
}
\end{figure*}

\begin{figure*}
\includegraphics[width=\textwidth]{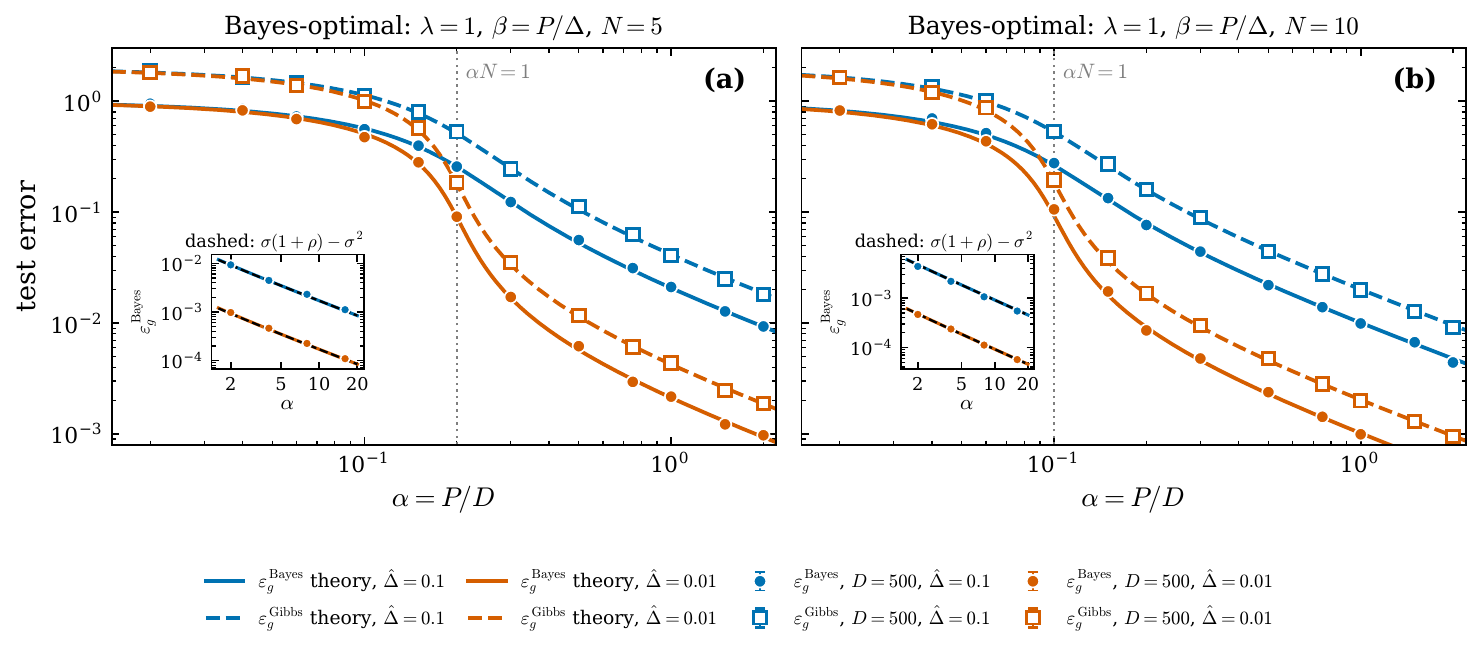}
\caption{\label{fig02}
Bayes optimal learning for finite-length context setting. Noisy labels of two different variances are considered. The insets show the theoretically predicted scaling laws compared with realistic simulations. 
}
\end{figure*}

As $\alpha N\to1^+$ the root of $\Upsilon(\chi)=1$ diverges. Hence, $\Upsilon\simeq\alpha N-\frac{\alpha h}{\theta\chi}$, where $h=\sum_cm_c/c$, so $\chi\simeq\frac{\alpha h}{\theta(\alpha N-1)}$, and $\varepsilon_g^{\rm Bayes}=\tau\chi\simeq\hat\Delta\frac{\alpha h}{\alpha N-1}$, indicating a divergence at $\a N=1$, yet cured by any $\lambda>0$.

Finally, we consider $b\to\infty$. $b\to\infty$ at fixed $\alpha$ means $\theta\to\infty$. The state equation implies that 
$\chi=\frac{1-\alpha N}\lambda+\frac{\alpha h}{\theta(1-\alpha N)}+O(\theta^{-2})$, provided that $\a N<1$. We then get
$Q_1=(1-\alpha N)+\hat\Delta\frac{\alpha h}{1-\alpha N}$~\cite{SM}. This leads to the noiseless Bayes error $(1-\a N)_{+}$. If $\a N>1$, then $\chi\to0$ with $u=\theta\chi$ finite. The state equation $\lambda\chi+\alpha\sum_cm_c\frac{cu}{1+cu}=1$ gives $u\to u_*$. Also $Q_1^{(0)}\to0$, so $Q_1=\tau\chi=\tau u_*/\theta=\hat\Delta u_*$. When $\alpha N\to1^+$, one obtains $u_*\simeq\frac{\alpha h}{\alpha N-1}$. Therefore, our theory predicts a double descent for the Bayes test error---a symmetric divergence at $\a N=1$.
A careful inspection reveals that for every $\alpha N>1$, the $b\to\infty$ Bayes error \emph{equals} the ridgeless one exactly~\cite{SM}, because the posterior mean depends only on $\lambda/\beta$. Moreover, the Gibbs error coincides with the Bayes error as for $\a N>1$ the thermal part decays as $\chi\simeq\a(N+1)u_*/b$.

  \begin{figure*}
\includegraphics[width=\textwidth]{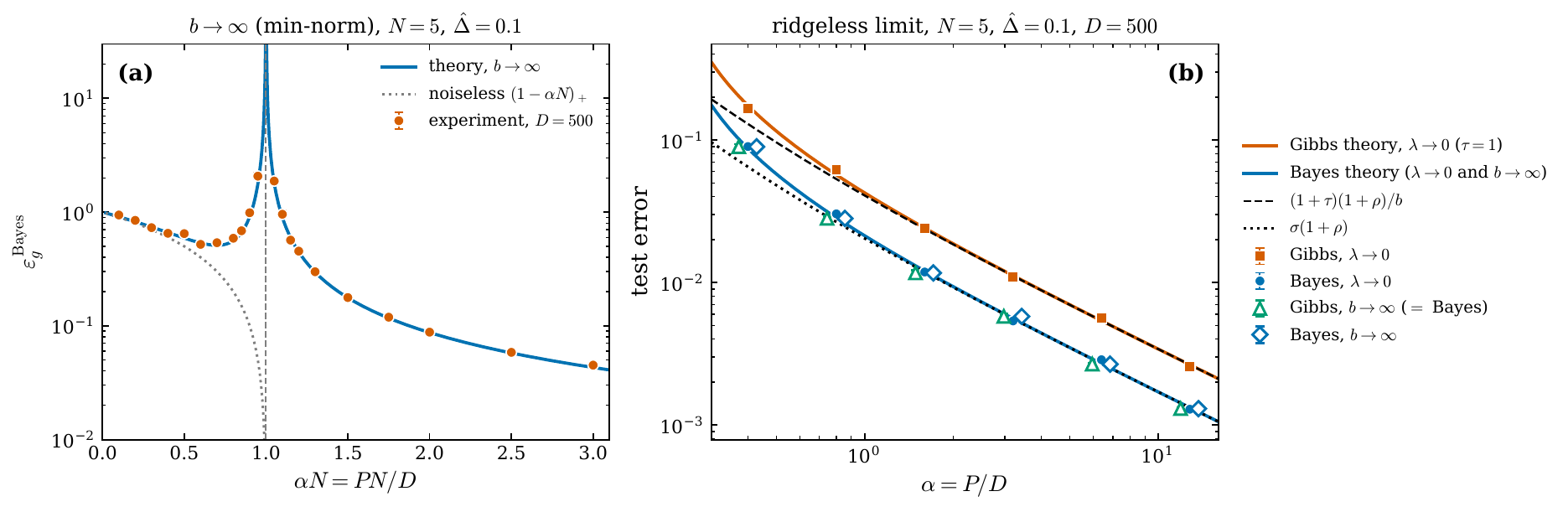}
\caption{\label{fig03}
Learning curves for zero-temperature and ridgeless learning. Finite-length context is considered. (a) Double descent phenomenon for zero-temperature learning. (b) The scaling laws for 
ridgeless learning. The zero-temperature learning is also compared.
}
\end{figure*}

\textit{Extensive context length.}---
In this case, $\rho=\zeta/(\a D)\to0$, and thus $\beta J=bI+O(\sqrt{\rho})$ in the bulk. Then $H\to(\lambda+b)I$. The errors are thus exactly 
those of \emph{deterministic ridge shrinkage}:
\begin{equation}
\chi=\frac1{\lambda+b},\qquad Q_1^{(0)}=\frac{\lambda^2}{(\lambda+b)^2},\qquad Q_1=\frac{\lambda^2+\tau b}{(\lambda+b)^2}.
\end{equation}
Therefore, we have $\varepsilon_g^{\rm Bayes}=\frac{1}{1+b}$ in the Bayesian optimal setting ($\tau=1$), while $\varepsilon_g^{\rm Bayes}=\frac\tau b$ in the ridgeless limit. We would not discuss a finite $P$ case, which becomes much more involved as the state equation must be modified.

\textit{Results.}---
We first consider noiseless learning without label noise. As shown in Fig.~\ref{fig01}, both test and training errors are well predicted by our theory. The Bayes error can be lower than the Gibbs training error, as the crossover occurs around $\a N=1$ where $Q_1=1/3$ in the current parameter setting. The training error grows with $\a$, which was also observed in standard perceptron learning~\cite{Haim-1992}. As expected, the Bayes error is lower than the Gibbs one.
In Fig.~\ref{fig02}, we study Bayesian optimal learning curves, where the Nishimori identity holds, and the Gibbs error is always two times the Bayes error.
Moreover, the Bayes error bears a power-law scaling with increasing $\a$: $\epsilon_g^{\rm Bayes}\sim\frac{1}{\a}$ at leading order. The theory perfectly matches the finite-$D$ learning experiments (see Insets of Fig.~\ref{fig02}). $\a N=1$ marks a rapid falloff of the learning curve.

   \begin{figure*}
\includegraphics[width=\textwidth]{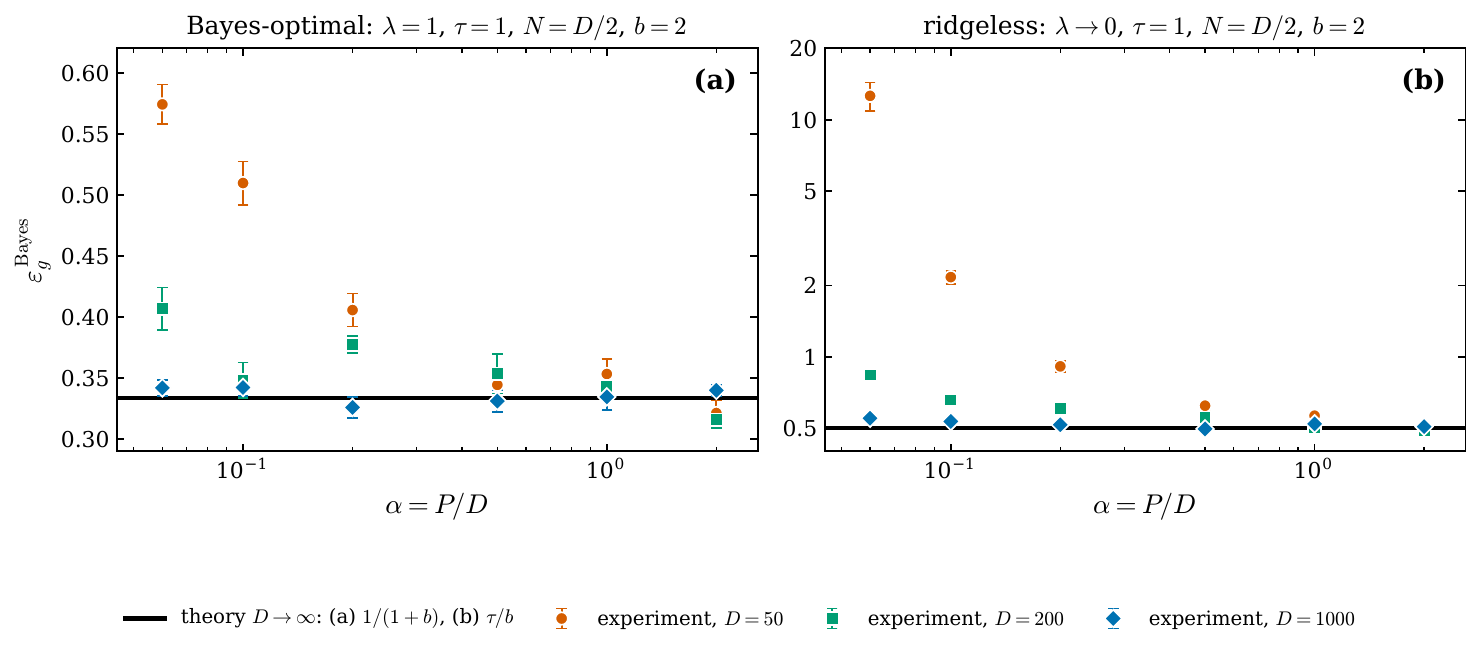}
\caption{\label{fig04}
Learning curves for extensive context length. (a) Bayes optimal learning. (b) ridgeless learning.
}
\end{figure*}

Next, we study the zero-temperature learning for finite context length and finite label noise. First, our theory predicts a double descent phenomenon where a peak occurs exactly at $\a N=1$ [Fig.~\ref{fig03} (a)]. The data load is inversely related to the context length. The Bayes error diverges symmetrically with $\kappa_{\pm}^{-1}$, where $\kappa_{\pm}=1-\a N$ or $\a N-1$. The learning dynamics around this interpolation point deserve further study. The double descent phenomenon was also empirically found in overparameterized deep neural networks and analytically studied in shallow neural networks~\cite{Belkin-2019,Montanari-2022}.  Note that the posterior mean depends only on $\lambda/\beta$~\cite{SM}. Therefore, both zero temperature and the ridgeless limit share the same asymptotic law for Bayes error when $\a\to\infty$, as shown both theoretically and experimentally in Fig.~\ref{fig03} (b).

  Finally, the learning curves of extensive context length are shown in Fig.~\ref{fig04}, which demonstrates that the decreasing behavior is only due to finite-$D$ effects, while in the thermodynamic limit, the Bayes error is a constant independent of $\a$, as predicted by our theory in the previous section.
 
\textit{Concluding remarks.}---
In this paper, we study the mechanics of next-token prediction using statistical mechanics of disordered systems and random matrix theory. We construct a teacher-student model to learn a high-dimensional token prediction given finite or extensive context. The test error scaling law for Bayesian optimal learning is derived as a power law, perfectly matching the experimental results. In zero-temperature learning, the Bayes test error displays a double descent phenomenon, suggesting that the data load and context length compete for the peak location at $\a N=1$. This double descent phenomenon deserves further study either in the learning dynamics aspect~\cite{Li-2026} or in the Landau theory framework~\cite{Kim-2026} to confirm the critical exponents derived in our theory. In the ridgeless limit, we obtain the same Bayes error as the zero-temperature learning for $\a N>1$. This study establishes an exact theory of next-token prediction, opening up possibilities for studying the neural scaling law for joint training.

\section{Acknowledgments}
This research was supported by the National Natural Science Foundation of China for
Grant number 12475045, and Guangdong Provincial Key Laboratory of Magnetoelectric Physics and Devices (No. 2026B1212170049). 
\section{AI use statement}
The results reported in this study were generated through extensive interaction with the powerful AI models GPT-6.1 sol and Claude Opus 5.5. These models were used for theoretical derivations and coding, guided by the authors. The authors take full responsibility for the scientific content. This manuscript reports our own thoughts on the problem defined by the senior author.





\end{document}